\pdfoutput=1
\documentclass[conference]{IEEEtran}
\usepackage{cite}
\usepackage{amsmath,amssymb,amsfonts}
\usepackage{algorithmic}
\usepackage{graphicx}
\usepackage{textcomp}
\def\BibTeX{{\rm B\kern-.05em{\sc i\kern-.025em b}\kern-.08em
    T\kern-.1667em\lower.7ex\hbox{E}\kern-.125emX}}

\usepackage[bookmarks=false,breaklinks=true]{hyperref}

\usepackage[utf8]{inputenc}
\usepackage{tikz}
\tikzstyle{startstop} = [rectangle, rounded corners, minimum width=3cm, minimum height=1cm,text centered, draw=black, fill=red!30]
\tikzstyle{io} = [trapezium, trapezium left angle=70, trapezium right angle=110, minimum width=3cm, minimum height=1cm, text centered, draw=black, fill=blue!30]
\tikzstyle{process} = [rectangle, minimum width=2cm, minimum height=1cm, text centered, text width=2.5cm, draw=black, fill=orange!30]
\tikzstyle{decision} = [diamond, minimum width=3cm, minimum height=1cm, text centered, draw=black, fill=green!30]
\tikzstyle{arrow} = [thick,->,>=stealth]
\usepackage{color,soul}

\newcommand{\vect}[1]{\boldsymbol{#1}}
\DeclareMathOperator*{\argmax}{arg\,max}

\usepackage{multirow}

\begin{document}

\title{Q-DeckRec: A Fast Deck Recommendation System for Collectible Card Games}

\author{
\IEEEauthorblockN{Zhengxing Chen}
\IEEEauthorblockA{\textit{College of Comp. and Info. Sci.} \\
\textit{Northeastern University}\\
Boston, USA \\
czxttkl@gmail.com}
\and
\IEEEauthorblockN{Christopher Amato}
\IEEEauthorblockA{\textit{College of Comp. and Info. Sci.} \\
\textit{Northeastern University}\\
Boston, USA \\
c.amato@northeastern.edu}
\and
\IEEEauthorblockN{Truong-Huy D. Nguyen}
\IEEEauthorblockA{\textit{Dept. of Comp. and Info. Sci.} \\
\textit{Fordham University}\\
Bronx, USA \\
tnguyen88@fordham.edu}
\and[\hfill\mbox{}\par\mbox{}\hfill]
\IEEEauthorblockN{Seth Cooper~}
\IEEEauthorblockA{\textit{College of Comp. and Info. Sci.} \\
\textit{Northeastern University}\\
Boston, USA \\
scooper@ccs.neu.edu}
\and
\IEEEauthorblockN{Yizhou Sun~}
\IEEEauthorblockA{\textit{Dept. of Comp. Sci.} \\
\textit{University of California, Los Angeles}\\
Los Angeles, USA \\
yzsun@cs.ucla.edu }
\and
\IEEEauthorblockN{Magy Seif El-Nasr~}
\IEEEauthorblockA{\textit{College of Comp. and Info. Sci.} \\
\textit{Northeastern University}\\
Boston, USA \\
m.seifel-nasr@northeastern.edu }
}

\maketitle

\begin{abstract}
Deck building is a crucial component in playing \textit{Collectible Card Games} (CCGs). The goal of deck building is to choose a fixed-sized subset of cards from a large card pool, so that they work well together in-game against specific opponents. Existing methods either lack flexibility to adapt to different opponents or require large computational resources, still making them unsuitable for any real-time or large-scale application. We propose a new deck recommendation system, named Q-DeckRec, which learns a deck search policy during a training phase and uses it to solve deck building problem instances. Our experimental results demonstrate Q-DeckRec requires less computational resources to build winning-effective decks after a training phase compared to several baseline methods.
\end{abstract}


\section{Introduction}\label{sec:introduction}
\textit{Collectible Card Games} (CCGs) have been popular since the 90s, evidenced by the large player base of these kinds of games. For instance, \textit{Magic: the Gathering} has more than 20 million players globally~\cite{guinnessmagic}, while an online free-to-play CCG \textit{Hearthstone} (Blizzard Inc.) reached a record of 40 million registered accounts in 2016~\cite{hearthstonepopular}.  

A CCG typically has hundreds to thousands of different cards, each of which supports specific in-game rules and effects. When playing CCGs, before each match, every player is asked to build a \textit{deck} comprising of a subset of all available cards. While in game, each player takes turns to draw cards from their respective deck and place them on the game board to fare (e.g., attack, counter-attack, cast spell, etc.) against their opponent cards.

In general, there is no single deck which can universally win against all other decks because CCGs often design cards with sophisticated synergistic and oppositional relationships. For example, in Hearthstone, there are two distinguished types of decks that counter each other in different phases of a match. An \textit{Aggro} deck, taking an aggressive approach, is built with cards capable of dealing damage to the opponents as quickly as possible. In contrast, a \textit{control} deck is the opposite archetype with cards which can survive long enough to triumph in the late game through powerful but expensive cards or complex combos.

The goal of deck building is to identify a set of cards which suits the player's own play style and effectively counters either an individual opponent or a group of opponents with specific play styles and decks. As deck building is regarded as a crucial part of game play, there exist many online forums and websites for players to discuss, analyze and test deck building strategies (e.g.,~\cite{hearthpwn,icyveins}).

A deck recommendation system for the purpose of deck building can benefit players and game developers in several ways. First, it can ease choices made by players in deck building. Players may also learn new strategies of deck building and practice their skills based on recommended decks. Second, such a system can be useful to increase player's engagement, by controlling match outcomes to keep players interested ~\cite{chen2017eomm,chen2015analytics}. Deploying a deck recommendation system in certain modes (e.g., a practice mode) could help re-engage players who are frustrated with the difficulty of building effective decks. Last, from a game developer's perspective, a deck recommendation system is also useful for debugging games. For example,  balancing the power of cards is an important topic in CCGs~\cite{ham2010rarity} or similar games~\cite{mahlmann2012evolving}. The game developer can use a deck recommendation system to check whether certain combinations of cards are powerful or weak.

As a deck is a combination of cards, deck building can be formulated as a combinatorial optimization problem (COP), which relates to finding an optimal solution (the most winning-effective deck) in a finite search space of all possible decks. Deck building has a large and complex solution space. For example, the number of all possible decks in our experiment setting, which selects 15 out of 312 cards, is $1.4 \times 10^{25}$. 

Previous works for deck building are mainly search algorithms, falling into two categories: heuristic searches and metaheuristic searches~\cite{birattari2009tuning}. Heuristic search methods decide which cards to include based on domain heuristics such as popularity and in-game resource curve~\cite{frankkarsten,willfancher,stiegler2016hearthstone}. However, heuristic methods require in-depth human knowledge and lack flexibility to adapt to different opponents. Another category is metaheuristic search, referring to high-level, problem-independent, approximate search strategies for tackling optimization problems~\cite{birattari2009tuning}. An example is to use a \textit{Genetic Algorithm} (GA)~\cite{holland1992adaptation} to evolve decks towards higher winning-effectiveness through repeated modifications and selections~\cite{garcia2016evolutionary,bjorke2017deckbuilding}. Although metaheuristic search algorithms do not require human knowledge to guide searches, they require a large computational cost \textit{for each deck building problem instance} because: (1) the search process requires a number of evaluations of candidate solutions; (2) the evaluation of a candidate solution's quality is computationally expensive, as this requires a large number of simulated matches with complicated in-game rules. 





An alternative view of solving the deck building problem is to treat it as a \textit{sequential decision making} problem~\cite{littman1996algorithms}.
Intuitively, a deck can be built by starting in some initial card configuration (i.e., state) and applying deck modification operators (such as adding, removing, or replacing an existing card) to move to new states. The goal is to end at a final state where the deck yields a high winning chance of winning against the opponent's deck. The key challenge is to decide which operator to apply in each state.  If a search policy (i.e., the mapping between states and operator choices) can be learned beforehand and simply followed while solving future problem instances, less computational resources will be needed compared to other methods requiring evaluating candidate solutions such as metaheuristic search. Such an idea is rooted in the paradigm of Reinforcement Learning (RL) algorithms concerned with learning a policy to maximize long-term rewards. In fact, leveraging RL to learn search policies for optimization problems has already been investigated in other domains, e.g.,~\cite{zhang2000solving,gaudel2010feature,zoph2016neural} but is novel for CCG deck building.

In this paper, we propose a deck recommendation system named \textit{Q-DeckRec} whose goal is to efficiently identify winning-effective decks against specific opponents. We first model the deck building problem as solving a COP by sequential decision making, then learn a search policy by leveraging an RL algorithm on a set of ``training" problem instances. The key idea is to generalize a search policy in order to find winning-effective decks and find them quickly for future problem instances. Thus, Q-DeckRec is suitable to deploy for large-scale or real-time application, e.g., an online CCG's backend to recommend winning-effective decks to a population of online players, a deck analysis website to serve hundreds of online visitors' deck building requests, or large-scale deck balancing tests.

The contributions of the paper are:
\begin{enumerate}
    \item we formulate the deck build problem as a combinatorial optimization problem (COP);
    \item we propose Q-DeckRec, an algorithm which learns a search policy for solving deck building problem instances quickly;
    \item we conduct experiments to demonstrate Q-DeckRec's suitability for large-scale or real-time application. The results show that after a training phase Q-DeckRec is able to build highly winning-effective decks within 9.63 seconds of CPU time, which is not achievable by other methods.
\end{enumerate}

The paper is structured as follows. Section~\ref{sec:previous_work} provides the general overview of related work. Section~\ref{sec:method} describes the details of problem formulation and the proposed algorithm. Section~\ref{sec:exp} and Section~\ref{sec:result} describe the details of our experiments and the results, respectively. Section~\ref{sec:limitations} discusses our limitations and future work. Section~\ref{sec:conclusion} concludes the study.

\section{Related Work}\label{sec:previous_work}

\subsection{Collectible Card Game Overview}
Although in-game rules may vary to some extent, we focus on those CCGs similar to Hearthstone because its pattern is common and the simulator we use for our experiments is also based on it. 

Each match is one-vs-one and turn-based. Each player starts with an amount of health and the goal is to destroy the opponent's health first. Each player is asked to construct a \textit{deck} of a fixed number of cards before the actual match. During a player's turn, he plays the cards drawn from his own deck as per their rules and limited by his resource. Cards can be mainly categorized as \textit{spells} and \textit{minions}. Spells are played, creating an effect on the battlefield, and then are discarded. Minions, on the other hand, stay in play, and can be used to attack the enemy or other minions. There usually exist several deck archetypes in a CCG and no single deck can triumph over others universally (see the example of Aggro and Control decks in Section~\ref{sec:introduction}). 


Although a player cannot know what cards constitute the deck of his opponent before the match, he could make predictions of opponent decks and propose his deck in advance to reflect his winning philosophies. In other deck building applications, such as recommending decks in a practice mode and deck balancing tests, opponent decks can also be assumed to be known at the time of deck building.

\subsection{Combinatorial Optimization}

An optimization problem consists of an objective function and a set of problem instances. Each problem instance is defined by a set of variables and a set of constraints among those variables. A \textit{candidate solution} to a problem instance is an assignment of values to the variables. A \textit{feasible solution} is a candidate solution that satisfies the set of constraints. An \textit{optimal solution} is a feasible solution that maximizes value of the objective function. A combinatorial optimization problem (COP) such as the traveling salesperson problem (TSP) is an optimization problem whose problem instances have finite numbers of candidate solutions. For many COPs, the number of candidate solutions is too large to exhaust in order to identify an optimal solution. 

It is not new to approximately solve COPs through various \textit{meta-heuristics}, i.e., high-level, problem-independent, approximate search strategies~\cite{birattari2009tuning}. In Genetic Algorithms (GA)~\cite{holland1992adaptation}, candidate solutions evolve towards better feasible solutions iteratively with \textit{mutation} and \textit{crossover} operators. In each generation, the \textit{fitness value} of every candidate solution is evaluated; the fitness value is usually the value of the objective function in the optimization problem being solved. The more fit candidate solutions are stochastically selected and modified to form a new generation. Another genre is called the \textit{Cross-Entropy} (CE) method~\cite{rubinstein1999cross}. The central idea is that the probability of locating an optimal solution using naive random search is a rare-event probability. CE can be used to obtain a new sampling distribution so that the rare-event is more likely to occur. Sampling from the new distribution will result in near-optimal solutions.

All the metaheuristic algorithms introduced so far are \textit{non-learning} search algorithms with one inherent disadvantage: \textit{each time a new problem instance arises} they require a number of objective function evalutions until a sufficiently high-quality feasible solution is found. Indeed, the search process is independent between different problem instances and does not generalize a search policy which could be simply followed without objective function evaluations. If objective function evaluation is computationally expensive, non-learning search algorithms would be inefficient to solve multiple problem instances of the same COP. 


Naturally, researchers are motivated to design algorithms to learn search policies for solving optimization problem instances~\cite{zoph2016neural,li2017learning,chenlearning,zhang2000solving}. These algorithms lie in a broader term known as ``meta-learning"~\cite{lemke2015metalearning,brazdil2008metalearning,vilalta2002perspective} or ``learning to learn"~\cite{thrun2012learning}. The learning of search policies relies on viewing the optimization process as conducting sequential decision making~\cite{littman1996algorithms} by an optimizer agent. The optimizer agent starts in some initial state and consecutively applies operators to move to new states. The goal is to end at a final state where a high-quality feasible solution can be extracted. We call the mapping between states and operator choices as the \textit{search policy}. If we additionally define a transition function and a reward function, we can formulate the optimization process as a \textit{Markov Decision Process} (MDP)~\cite{bellman1957markovian}. The optimal policy that maximizes long-term rewards can be learned or approximated by leveraging reinforcement learning (RL) algorithms~\cite{sutton1998reinforcement}. The key is to properly design the MDP, especially the reward function, such that the learned policy can guide the optimizer agent quickly towards high-quality feasible solutions. For example, Zhang and Dietterich applied an RL algorithm $TD(\lambda)$ to obtain the search policy for solving NASA space shuttle scheduling problem instances~\cite{zhang2000solving}. Their results show that the learned search policy is more effective in the ratio of solution quality vs. CPU time than the best known non-learning search algorithm on test problem instances. Bello et. al show that a search policy parameterized as a special structure of neural network can be trained and used to solve unseen instances of TSP~\cite{zoph2016neural}. In Section~\ref{sec:method}, we will show that under certain assumptions, a deck building problem can be formulated as a COP and a search policy learned by RL  on ``training" problem instances will quickly guide building winning-effective decks on future problem instances. 

Besides metaheuristic and RL algorithms, problem-dependent heuristics can be used to search solutions when COPs have exploitable characteristics. However, designing heuristics is a labor intensive job. Researchers have also attempted to use supervised learning models to learn the mapping from problem instances to optimal solutions (e.g., \cite{vinyals2015pointer}). Optimal or approximated optimal solutions need to be calculated by some solver in advance in order to provide training supervised signals. One difficulty is to design special model architectures to cope with discrete nature and constraints of COPs. For instance, in the TSP, the outputs should be constrained to sequences with no duplicated cities~\cite{vinyals2015pointer}. 

\subsection{Deck Building}

Ideas proposed for deck building mainly fall into two categories: heuristics and metaheuristic searches. First, some heuristic methods decide which cards to include based on the popularity of cards from  historical data~\cite{frankkarsten,willfancher}. The underlying intuition is that popularly favored cards are very likely to be strong ones. Stiegler \textit{et al.} propose a utility system to search deck with more types of game-specific heuristics besides card popularity, including mana curve, strategic parameters, cost effectiveness and card synergies~\cite{stiegler2016hearthstone}. However, all heuristic methods require intensive human knowledge, lack  flexibility to adapt to different opponent decks, and are not easy to transfer to other games intelligently. As far as we know, \textit{Genetic Algorithm} (GA)~\cite{holland1992adaptation} has been the only metaheuristic search algorithm for deck building~\cite{garcia2016evolutionary,bjorke2017deckbuilding}. In their works, the fitness value is the average win rate of a candidate deck against a group of opponent decks while AI bots are used as a proxy for human play. However, we note in their results that a single run of GA for a particular deck building problem instance took hours or days to reach a winning-effective deck~\cite{garcia2016evolutionary,bjorke2017deckbuilding}. This is because each fitness evaluation requires obtaining a win rate based on a number of simulated matches. The complicated in-game rules also make the simulation computationally expensive. Therefore, non-learning search algorithms like GA are not practical for any large-scale or real-time deck recommendation task.





\section{Methodologies}\label{sec:method}
In this section we will formally describe how the deck building problem can be cast as a COP. We will then proceed to presenting our proposed solution, \textit{Q-DeckRec}, which solves the deck building problem from a sequential decision making perspective. In this paper, we focus on building winning-effective decks against specific individual opponents. We will leave the discussion about building decks against a group of opponents in Section~\ref{sec:conclusion}. 

\subsection{Problem Formulation}
Deck building can be formulated as a combinatorial optimization problem (COP). 
Suppose the goal is to build decks as a subset of size $D$ among a total of $N$ cards with $N > D $ (usually $N$ is several times larger than $D$). A deck can be represented as a binary vector of length $N$, $ \vect{x} \in \mathbb{Z}_2^N$, whose components of 1's correspond to the cards included in the deck and 0's otherwise. Since a deck has a fixed size of cards, we have $\|\vect{x}\|_1=D$. We use $\vect{x}_p$ and $\vect{x}_o$ to differentiate the deck of the player and his opponent. We use $\mathcal{A}_p$ and $\mathcal{A}_o$ to capture the play styles of the player and his opponent. $\mathcal{A}_p$ and $\mathcal{A}_o$ are play style-specific simulators that decide which cards to issue given a game instance, henceforth referred to as the \textit{AI proxies} (artificial intelligence) of respective play styles. The evaluation function $f(\cdot)$ is defined as $f(\vect{x}_p; \vect{x}_o, \mathcal{A}_p, \mathcal{A}_o)$, which returns the winning probability of the player using $\vect{x}_p$ against the opponent using $\vect{x}_o$, with their play styles following $\mathcal{A}_p$ and $\mathcal{A}_o$ respectively. The objective function of the deck building problem is formulated as:
\begin{align}
\begin{split}
\argmax_{\vect{x}_p} & \;\; f(\vect{x}_p; \vect{x}_o, \mathcal{A}_p, \mathcal{A}_o) \\
\text{subject to } \;\;&\vect{x}_p \in \mathbb{Z}_2^N, \vect{x}_o \in \mathbb{Z}_2^N, \\
&\|\vect{x}_p\|_1=\|\vect{x}_o\|_1=D
\end{split}
\label{eqn:obj}
\end{align}
the solution of which is denoted as $\vect{x}_p^*$.

Note that $f(\cdot)$ is a black-box function. We do not have the closed-form expression of $f(\cdot)$, but can approximate its value by simulating $\mathcal{A}_p$ and $\mathcal{A}_o$ playing against each other for a number of matches. In practice, a sufficient number of matches need to be simulated to get stable win rate estimation. Since the simulation needs to apply numerous rules of the game on each move, this is a computationally demanding operation. Each evaluation of $f(\cdot)$ was found to take non-negligible time in the order of seconds on a very powerful server machine (see Section~\ref{sec:exp}). Therefore, the brute-force approach is almost infeasible to apply, as it needs evaluate an exponential number of $\vect{x}_p$ configurations, i.e., ${N\choose D}=N!/D!/(N-D)!=O(N^D)$. For example, in our experimental setting where $N=312, D=15$, it would need to exhaust around $1.4 \times 10^{25}$ possibilities.

After examining potential usage scenarios, we find one assumption that could be made and exploited. Although different problem instances may use different $\mathcal{A}_p$ and $\mathcal{A}_o$, we assume that $\mathcal{A}_p$ and $\mathcal{A}_o$ come from a pool of AI proxies pre-trained by a deck recommendation system. For example, each AI proxy from the pool represents a specific play style archetype such as ``aggressive" or ``conservative". Under this assumption, each problem instance consists of $\vect{x}_o$ which may vary, and $\mathcal{A}_p$ and $\mathcal{A}_o$ which have been available. Therefore, there might exist deck building patterns which can be generalized. For example, if certain $\mathcal{A}_p$ is good at using Card A to counter certain $\mathcal{A}_o$, then Card A tends to appear in the optimal solution of many problem instances with the two AI proxies as the input.

In the rest of the paper, we will assume we deal with deck building problem instances of Eqn.~\ref{eqn:obj} under a specific pair of $\mathcal{A}_p$ and $\mathcal{A}_o$. All the methodologies will be invariant for other pairs of AI proxies.

\subsection{Q-DeckRec}




We propose to delegate the problem of generalizing deck building patterns as a problem of generalizing a search policy in a Markov Decision Process (MDP)~\cite{bellman1957markovian} environment, where an agent naviagates in the state space to search for the most winning-effective deck. 

In the MDP, a state $s \in \mathcal{S}$ consists of a unique feasible solution $\vect{x}_p$, together with $\vect{x}_o$ and a step counter $t$ as complement information, i.e., $s=\{\vect{x}_p, \vect{x}_o, t\}$. An action $a \in \mathcal{A}$ is defined as a card replacement to modify the current deck $\vect{x}_p$. An action replaces exactly one card in the deck $\vect{x}_p$ with another card not included currently. One special action is to keep the current deck as unmodified. Given the actions we define, the transitions between states $T: \mathcal{S} \times \mathcal{A} \rightarrow \mathcal{S}$ are always deterministic. One state applied by an action will transit to only one next state, reflecting the corresponding card modification, denoted as $\{\vect{x}^{(t)}_p, \vect{x}_o, t\}, a \rightarrow \{\vect{x}^{(t+1)}_p, \vect{x}_o, t+1\}$. The deck search starts from a random initial state $s_0=\{\vect{x}_p^{(0)}, \vect{x}_o, 0\}$ and is limited to take exact $D$ actions in one episode. We denote the states within one episode as $s_0, s_1, \cdots, s_D$. We limit the length of the horizon to be $D$ because at most we need to replace all the cards in $\vect{x}_p^{(0)}$ to reach the optimal deck $\vect{x}_p^*$.

The problem remains as how to design the reward function $R: \mathcal{S} \times \mathcal{A} \rightarrow \mathbb{R}$. In the MDP, the optimal policy is the one which maximizes a defined long-term reward criterion. The  key is to properly design the reward function and long-term reward criterion, such that the optimal policy is indeed the desired search policy which can lead to winning-effective decks from any state. 

The long-term reward criterion defines the goal of reinforcement learning. It should encourage the optimal policy to search in the direction of winning-effective decks. We propose the following long-term reward criterion for each episode:
\begin{align}
R &=\sum_{t=0}^{D-1} r_t,
\end{align}
where $r_t$ is the reward function over each transition. Specifically, we define $r_t$ as the win rate between the opponent deck and the modified deck after step $t$ with exponential amplification:
\begin{align}
    r_t=exp(b \cdot f(\vect{x}^{(t+1)}_p; \vect{x}_o, \mathcal{A}_p, \mathcal{A}_o)),
\end{align}
where $b$ is a positive constant to adjust the extent of amplification. We choose this reward function over $r_t=f(\vect{x}^{(t+1)}_p; \vect{x}_o, \mathcal{A}_p, \mathcal{A}_o)$ in order to amplify the difference between strong and weak decks. Although the goal of deck building is to land on ${s_D=\{\vect{x}_p^{(D)}, \vect{x}_o, D\}}$ with $f(\vect{x}_p^{(D)}; \vect{x}_o, \mathcal{A}_p, \mathcal{A}_o)$ as high as possible, the cumulative sum of win rates provides more reward signals along the search than merely optimizing ${R=r_D}$. This shape of reward has also been used in previous optimization problems based on sequential decision making~\cite{andrychowicz2016learning,chenlearning}. Since we model each episode with finite horizons, we ignore the conventional reward discount factor $\gamma$ in the definition of $R$, which is a mathematical trick to help the convergence of RL learning in MDPs with infinite horizons. 

The optimal policy can be obtained by always selecting the action with the highest optimal state-action value at each state:
\begin{align}
    \pi^*(s) = \argmax_a Q^*(s,a),  s = s_0, \cdots, s_{D-1},
\label{eqn:optpolicy}
\end{align}
where $Q^*(s,a)$ is defined as the best state-action value function among all possible policies:
\begin{align}
Q^\pi(s,a)=\mathbb{E}[\sum_{i=t}^{D} r_i | s_t=s, a_t=a, \pi] 
\end{align}
\begin{align}
Q^*(s,a)=\max_\pi Q^\pi(s,a) 
\end{align}
The intuition behind $Q^*(s,a)$ is that it measures how promising applying the modification on the current deck would lead to the most winning effective deck. Following $\pi^*(s)$ would generate a series of modifications that faithfully build the optimal deck. 

We propose to use a Reinforcement Learning (RL) algorithm, Q-Learning~\cite{watkins1992q}, to learn $Q^*(s,a)$ iteratively through observation tuples $(s,a,r,s')$. The simplest implementation of Q-Learning is a look-up table and a learning rate $0 < \alpha \leq 1$, with the update rule as: 
\begin{align}
\hat{Q}(s,a) = (1 - \alpha) \hat{Q}(s,a) + \alpha (r + \max_{a'} \hat{Q}(s', a')) 
\label{eqn:qsa}
\end{align}

Theory implies that if each action is tried in each state an infinite number of times and the magnitude of $\alpha$ meets certain criteria, then $\hat{Q}$ converges to $Q^*$~\cite{bertsekas1989parallel}. However, our problem has a huge state space hence it is not possible to maintain a look-up table for all combinations of states and actions. Instead, we resort to Multi-Layer Perceptron (MLP) with parameters $\theta$ as a function approximator: $Q_\theta(s,a)$ learns to approximate the mapping of the feature representation of the state-action pair, $\mathcal{F}(s,a)$, to the optimal state-action value, $Q^*(s,a)$. More specifically, we use an MLP architecture with one input layer, one hidden layer and one output layer. Without requiring any prior domain knowledge, we simply let $\mathcal{F}(s,a)=s'$. Therefore, the input layer takes as input a state representation $s'$, which has ${2\cdot N+1}$ dimensions. The output layer outputs a real value  representing the predicted $Q^*(s,a)$. The exact specifications can be seen in Section~\ref{sec:exp}. The update rule of $\theta$ is in a gradient descent fashion towards reducing so-called TD-error $\delta$:

\begin{align}
\delta := r + \max_{a'} Q_{\theta}(s', a') - Q_{\theta}(s,a)
\label{eqn:tderror}
\end{align}

\begin{align}
\theta \leftarrow \theta + \alpha \cdot \delta \cdot \nabla Q_\theta(s,a)
\label{eqn:updaterule}
\end{align}

To learn $\theta$, we need to collect  observation tuples $(s,a,r,s')$ through solving "training" problem instances. Solving a training problem instance is to let Q-DeckRec take actions $D$ times based on the current $Q_\theta(s,a)$ function in an episode. In order to generalize $Q_\theta(s,a)$ to various states, we initialize both $\vect{x}_o$ and $\vect{x}_p^{(0)}$ in $s_0$ randomly at the beginning of each episode. An $\epsilon$-greedy policy is used during the training, with $\epsilon$ slowly decreasing as the learning proceeds. The policy has $\epsilon$ probability to choose non-optimal actions in the hope to escape any local optimum and discover better policies. Also, we use prioritized experience replay~\cite{schaul2015prioritized} to improve sample efficiency. Past experiences will be weighted according to the absolute value of $\delta$. High TD-error associated experiences will be more likely to be sampled for MLP parameter learning. 

The training phase of Q-DeckRec can be summarized as follows. At the beginning of each training episode, both $\vect{x}_o$ and $\vect{x}_p^{(0)}$ are randomly generated. Q-DeckRec decides how to ``navigate" through states by $\epsilon$-greedy policy and $Q_\theta(s,a)$ in $D$ steps. All the $D$ transitions are stored into the prioritized experience replay pool. Following that, $m$ previous observation tuples $(s, a, r, s')$ are sampled from the prioritized experience replay as a learning batch for updating $\theta$ as described in Eqn.~\ref{eqn:tderror}~and~\ref{eqn:updaterule}. The loop continues after a new training episode is initiated. The training will be terminated after a time limit is reached. 

After training, $Q_\theta(s,a)$ will become fixed. When solving a future problem instance, Q-DeckRec can start from $s_0$ with a random $\vect{x}_p^{(0)}$ and follows $\pi^*$ as in Eqn.~\ref{eqn:optpolicy} in $D$ steps. No call of $f(\cdot)$ will be needed during the search. As a comparison, non-learning search algorithms such as Genetic Algorithm require calling $f(
\cdot)$ multiple times in order to evaluate fitness values for each problem instance~\cite{garcia2016evolutionary,bjorke2017deckbuilding}, while calling $f(\cdot)$ would take computational resources much heavier than calculating $Q_\theta(s,a)$. Therefore, Q-DeckRec has its superior suitability for large-scale or real-time application.

\section{Experiment Setup}\label{sec:exp}
To verify our method and compare with other methods, we test on an open-sourced CCG simulator \textit{MetaStone}\footnote{\url{https://github.com/demilich1/metastone}}, which is  based on the popular online digital CCG \textit{Hearthstone} (Blizzard Entertainment, Inc.). All experiments run on a powerful server with Intel E5 2680 CPU’s @ 2.40 GHz (56 logical CPU cores). Parallelization is implemented in three places: (1) linear algebra operations used in the MLP in Q-DeckRec; (2) match simulations evenly spread on all cores when evaluating $f(\cdot)$; (3) random deck sampling from a baseline based on Monte Carlo simulations (introduced later). Each call of $f(\cdot)$ returns a win rate based on 300 simulated matches, which on average takes 5 seconds and has around 5\% standard deviation in the win rate evaluation. 

We make a few decisions in setting up our experiments. We expect the experiment results can generalize under other settings. First, we use the same AI proxy to represent both $\mathcal{A}_p$ and $\mathcal{A}_o$. The AI proxy is provided by the simulator and is called \textit{GreedyOptimizeMove}. It decides the best action by evaluating each action's consequence according to a heuristic. We do not use other AI proxies based on tree search methods because they take much longer per match. Second, we assume both players are from a specific in-game character class called \textit{Warriors}. The total number of available cards to Warriors is 312. Third, while in the real game certain cards can have at most one copy and all other cards can have at most two copies in the deck, we impose that every included card has two copies. This reduces our search space size for the test purpose and also follows the postulation that having two copies for every card makes the deck performance more reliable~\cite{garcia2016evolutionary}. As a result, although the deck size is 30, the number of cards to be selected is 15. In summary, we have $N=312, D=15$ when optimizing Eqn.~\ref{eqn:obj}.

We set up Q-DeckRec as follows. The underlying MLP has one hidden layer and one output layer. The hidden layer consists of 1000 rectified linear units (ReLU). The output layer is a single unit which outputs a weighted sum from the activation values of the hidden layer. $\epsilon$ in the $\epsilon$-greedy policy starts at $1$ and decreases $0.0005$ per training episode until it reaches $0.2$. The size of a learning batch, $m$, is set at 64. For the prioritized experience replay~\cite{schaul2015prioritized}, the exponent $\alpha$ is set at 0.6, the exponent $\beta$ is linearly annealed from $\beta_0=0$ to $1$ with step $1e^{-5}$. The capacity of the experience pool is 100K. The constant $b$ in the reward function is set as 10. All the hyperparameters are chosen empirically without fine-tuning due to large computational resources required.  

We compare Q-DeckRec with a Genetic Algorithm (GA), the method used in previous works for deck building~\cite{garcia2016evolutionary,bjorke2017deckbuilding}. We implement GA with an open source library \textit{DEAP}\footnote{\url{https://github.com/DEAP/deap}}. An individual is a candidate deck $\vect{x}_p$. The fitness value is $f(\vect{x}_p; \vect{x}_o, \mathcal{A}_p, \mathcal{A}_o)$. The mutation and crossover functions are customized to maintain the validity of individuals, similarly to what was adopted in~\cite{bjorke2017deckbuilding}. Specifically, mutation is swapping one card in the deck with one not in the deck and crossover randomly exchanges cards not overlapped by the two decks. The population size of each generation is 10, with the mutation probability and the crossover probability both set as 0.2. Individual selection is based on a commonly used selection mechanism called \textit{tournament} of size 3. 

We also design an ad-hoc baseline which, like Q-DeckRec, requires a learning phase and does not require calling $f(\cdot)$ for solving future problem instances. The baseline conducts Monte Carlo (MC) simulations using a win rate predictor $\hat{f}(\cdot)$ to locate a solution. We first train a supervised learning model to approximate $f(\cdot)$. The training data are randomly generated pairs of decks represented as binary vectors. The labels are the evaluated win rates based on $f(\cdot)$. We choose to train an MLP with the same architecture as in Q-DeckRec. Given the same input, $\hat{f}(\cdot)$ would output faster than $f(\cdot)$ because the former does not need a real match simulation. When solving a future problem instance with opponent deck $\vect{x}_o$, we run MC simulations according to:

\begin{align}
\argmax_{\vect{x}_p \in \mathcal{X}_p} \hat{f}(\vect{x}_p, \vect{x}_o; \mathcal{A}_p, \mathcal{A}_o),
\label{eqn:mcmaxobj}
\end{align}
where $\mathcal{X}_p$ is a set of randomly generated decks. We denote the size of $\mathcal{X}_p$ as $X$. A larger $X$ means more thorough sampling.

In the experiments, we do not include any heuristic search method because we focus on algorithmic deck recommendation systems requiring minimal human knowledge involved. Besides GA, we do not include other metaheuristic search methods; similarly to GA, they all require calling the win rate evaluation function $f(\cdot)$ a number of times while solving each problem instance. We do not include  supervised learning models which directly learns the mapping from problem instances to optimal solutions because this requires designing a specific model architecture to cope with the characteristics of the deck recommendation problem (e.g., outputs are constrained to contain $K$ cards), which has not been studied before and requires non-trivial extra works.

Different wall time (i.e., real elapsed time) limits are imposed as the termination condition for both Q-DeckRec training and one run of GA. In this way, we can compare how long Q-DeckRec training and a GA run would take to reach similar performances. Wall time limits are chosen empirically based on observations in preliminary experiments and our limited computational resources. For GA, we try wall time limits as 10, 15, 20 and 25 minutes because performances often plateau after 20 minutes (as evidenced in the result section). Since we do not have the optimal solutions for test problem instances, we reference the solutions from 25-minute GA searches as approximated ground truths. For Q-DeckRec training, we test one, two and three days as the wall time limit. As will be shown in the result section, Q-DeckRec after three-day training can already reach the same optimality level as GA with 25 minute search. For the MC-simulation method, we use a training data set collected in three days and test ${X=67, 670, 6.7K, 67K, 670K}$ and $6700K$. Note that $67K$ is around the same number Q-DeckRec calls its learned function approximator $Q_\theta(s,a)$ for solving a test problem instance\footnote{As in Eqn.~\ref{eqn:optpolicy}, each optimal action is decided after calculating the state-action values of all possible actions ($(N-D)\cdot D + 1$) and we need to take $D$ actions per episode. When $N=312$ and $D=15$, the total number of state-action value evaluations is 66840.} whereas higher values of $X$ than $6700K$ would require too large computational resources to be practical for large-scale or real-time application. 

In the rest of the paper, we will denote an algorithm as a specific approach ($GA$, $Q\textnormal{-}DeckRec$, or $MC$) plus an associated parameter. For example, $GA_{20min}$ and $MC_{670K}$ are two algorithms. So are $Q\textnormal{-}DeckRec_{1day}$ and $Q\textnormal{-}DeckRec_{2days}$.   

We generate 20 test problem instances for evaluating all algorithms. In our preliminary experiments where test problem instances are randomly generated, we often find GA only needs less than 100 calls of $f(\cdot)$ to identify decks with 100\% win rate. This is because randomly generated $\vect{x}_o$ barely has any effective card synergy and can be easily beaten by a mediocre deck. In real-world applications, we believe it is more demanding to build winning-effective decks against competitive decks rather than random decks. In order to generate competitive opponent decks as test problem instances, we adopt a sequential manner as follows. We sample a deck $\vect{x}$ from the outputs of all algorithms for the last test problem instance, where the sampling distribution is weighted by $f(\vect{x};\vect{x}_o, \mathcal{A}_p, \mathcal{A}_o)$. We then use $\vect{x}$ as the input $\vect{x}_o$ for the next problem instance. The first test problem instance is obtained after 10 preliminary runs.

For a problem instance and an algorithm, the win rate of the returned $\vect{x}_p$ vs. the input $\vect{x}_o$ is considered as the result performance. We run each algorithm on each test problem instance 10 times. Each run is associated with a random seed, which controls the initialization of $\vect{x}_p^{(0)}$ in $s_0$ in Q-DeckRec, and the randomness in evolution behaviors in GA. Then, we use the median of the 10 runs as the performance for the algorithm on the problem instance. To measure the significance of the differences for each pair of algorithms, we also conduct a two-tailed paired Welch's t-tests with a confidence level 0.01 over all test instances. The null hypothesis is that the mean difference between the paired algorithms' win rates is zero.

In order to give a complete view of resource usage, we record both wall time and CPU time each algorithm takes to solve a test problem instance.

\begin{table}
\caption{Results of Genetic Algorithm Approach} 
\centering 
\begin{tabular}{c c c} 
\hline
 \begin{tabular}{@{}c@{}} Search Time \\ Wall Time / CPU Time\end{tabular} &  Func. Calls &  Win Rate
\\ 
\hline 
10 min / 6.2 hr & 110  & 0.61 \\
15 min / 9.3 hr & 189 & 0.86 \\
20 min / 12.8 hr & 267 & 0.93 \\
25 min / 15.9 hr & 315 & 0.94 \\
\hline 
\end{tabular}
\label{tab:ga}
\end{table}

\begin{table}
\caption{Results of Q-DeckRec} 
\centering 
\begin{tabular}{c c c c} 
\hline
 \begin{tabular}{@{}c@{}} Training \\ Wall Time\end{tabular} & \begin{tabular}{@{}c@{}} Search Time \\ Wall / CPU Time\end{tabular} & Func. Calls &  Win Rate \\
\hline
1 day & & 20K & 0.64 \\
2 days & 0.38 sec / 9.63 sec & 41K & 0.88 \\ 
3 days &  & 62K & 0.93  \\ 
\hline

\hline 
\end{tabular}
\label{tab:qdeckrec}
\end{table}

\begin{table}
\caption{Performances of MC simulation approach} 
\centering 
\begin{tabular}{c c c} 
\hline
$X$ (Number of Samples) &  \begin{tabular}{@{}c@{}} Search Time \\ Wall Time / CPU Time\end{tabular} &  Win Rate
\\
\hline 
$67$ & 0.01 sec / 0.18 sec & 0.48  \\
$670$ & 0.03 sec / 0.91 sec & 0.64 \\
$6.7K$ & 0.05 sec / 2.16 sec & 0.75 \\
$67K$ & 0.45 sec / 8.45 sec & 0.84 \\
$670K$ & 4.90 sec / 97.60 sec & 0.82 \\
$6700K$ & 36.06 sec / 1031.81 sec & 0.77 \\
\hline 
\end{tabular}
\label{tab:mc}
\end{table}

\section{Results and Discussion}\label{sec:result}

The performances of the three kinds of methods are reported in Table~\ref{tab:ga},~\ref{tab:qdeckrec} and~\ref{tab:mc}. All the reported numbers are the mean results over the 20 test problem instances. As stated, the result for each test problem instance is the median of 10 runs. Also, we find that all pairwise comparisons on the win rate are significant, \textit{except}: (1) $GA_{20min}$ vs. $GA_{25min}$ (2) $Q\textnormal{-}DeckRec_{3days}$ vs. $GA_{20min}$ (3) $Q\textnormal{-}DeckRec_{3days}$ vs. $GA_{25min}$.

First, we observe that the performances of GA and Q-DeckRec improve as the wall time limits increase in our test ranges. This meets our expectation because approximate COP solvers are supposed to get better solutions if using more computational resources. However, longer wall time limits than 20 minutes bring diminishing improvement in GA as we find there is no significant difference in the average win rate between $GA_{20min}$ and $GA_{25min}$. 

From Table~\ref{tab:ga}, we observe that GA calls the win rate evaluation function an increasing number of times as the wall time limit increases. As we stated, the win rate evaluation is  computationally expensive involving simulating 300 matches. Therefore, all GA algorithms require high CPU time in the order of hours. 


As shown in Table~\ref{tab:qdeckrec}, Q-DeckRec can solve deck building problem instances with as little computational cost as 9.63 seconds in CPU time. Meanwhile, Q-DeckRec after 3-day training can build decks as winning-effective as $GA_{25min}$ does, as evidenced by the non-significant difference between $Q\textnormal{-}DeckRec_{ 3days}$ vs. $GA_{25min}$. Therefore, from the CPU time perspective, Q-DeckRec is much efficient than GA ({9.63 sec $\ll$ 15.9 hr}) to solve a new problem instance because the computationally heavy match simulations have been "moved" to the training phase. This proves the merit of Q-DeckRec being a suitable deck recommendation system for large-scale or real-time application. 

The number of function calls is $62K$ during the training of $Q\textnormal{-}DeckRec_{3days}$. This means there are $62K$ state transitions generated from roughly $4K$ ($\approx 62K/15$) training episodes. Even if each of the $62K$ state transitions is unique, they still involve a tiny fraction of total states in our formulated state space. (The number of total states is the number of possible opponent decks times the number of possible player decks: ${N\choose D} \times {N\choose D} \approx 1.97 * 10^{50}$.) This shows that the MLP-based architecture is a well-chosen function approximator for generalizing state-action values.    


For the MC-simulation method, we first report Mean Squared Error (MSE) and $R^2$ of the learned supervised learning model. We evaluate them using a standard 10-fold cross validation. On training data, $MSE=0.005$ and $R^2=0.86$. On testing data, $MSE=0.008$ and $R^2=0.79$. To our surprise, from Table~\ref{tab:mc}, we find that the win rate does not monotonously increase as $X$ increases. The performance peaks at 0.84, which is significantly lower than $Q\textnormal{-}DeckRec_{3days}$. While debugging the method, we observe that the predicted win rate (the outcome of Eqn.~\ref{eqn:mcmaxobj}) monotonously increases as $X$ increases. We suspect that since the supervised learning model cannot perfectly predict the real win rate, deck samples inevitably contain outlier decks with spuriously high predicted win rates. These outlier decks "trick" the MC-simulation method to select them unfortunately. The results show that the approach of building winning-effective decks in a sequential way as in Q-DeckRec is more robust.

\section{Limitations and Future Works}\label{sec:limitations}
In order to help human players, Q-DeckRec relies on AI proxies which can accurately model players' play styles. The current used AI proxy is only based on a greedy heuristic rather than trained on human play traces. Training human-like AI proxies and integrating them to Q-DeckRec will be an important direction in our future works.

We can also improve sample efficiency in Q-DeckRec. Currently, a training episode starts with a random $s_0 = \{\vect{x}_p^{(0)}, \vect{x}_o, 0\}$. Were it generated from a card distribution learned from real matches, Q-DeckRec can focus on exploring in a smaller state space.

Next, as online CCGs often release patches to introduce new cards and modify existing cards' in-game effects, we would like to investigate how Q-DeckRec can transfer and update its knowledge without totally re-training the model~\cite{taylor2009transfer}.

Lastly, if the problem is extended to recommend winning-effective decks against a group of opponent decks $\{\vect{x}_{oi}\}^k_{i=1}$, there remains a question of how to design the feature representation of state-action pairs. Naive feature representations for the opponent deck group could be simply concatenating $\{\vect{x}_{oi}\}^k_{i=1}$. However this creates a large feature space which may not be efficient for learning. A more advanced feature representation may represent the opponent deck group in a continuous vector space, similar to word-embedding techniques from Natural Language Processing (NLP)~\cite{mikolov2013distributed}. We intend to investigate all of these in the future.




\section{Conclusions}\label{sec:conclusion}
In this paper, we propose a deck recommendation system named \textit{Q-DeckRec}, which is able to solve deck building problem instances in large-scale and real-time after a period of training and requires minimal domain knowledge. We design experiments that demonstrate the advantages.

\bibliographystyle{IEEEtran}
\bibliography{bib_game,bib_tech}

\end{document}